\documentclass[letterpaper]{article} 
\usepackage{aaai24}  
\usepackage{times}  
\usepackage{helvet}  
\usepackage{courier}  
\usepackage[hyphens]{url}  
\usepackage{graphicx} 
\usepackage{natbib}  
\usepackage{caption} 
\usepackage{algorithm}
\usepackage{algorithmicx}
\usepackage{algpseudocode}

\usepackage{newfloat}
\usepackage{listings}
\DeclareCaptionStyle{ruled}{labelfont=normalfont,labelsep=colon,strut=off} 
\floatstyle{ruled}
\newfloat{listing}{tb}{lst}{}
\floatname{listing}{Listing}
\usepackage{tabularx}
\usepackage{graphbox}
\usepackage{multirow}
\usepackage{booktabs}
\algrenewcommand\algorithmicrequire{\textbf{Dataset:}}
\algrenewcommand\algorithmicensure{\textbf{Base Model:}}

\title{Knowledge-tuning Large Language Models with Structured Medical Knowledge Bases for Reliable Response Generation in Chinese}
\author {
    Haochun Wang, Sendong Zhao, Zewen Qiang, Zijian Li, Nuwa Xi, Yanrui Du, MuZhen Cai, Haoqiang Guo, Yuhan Chen, Haoming Xu, Bing Qin, Ting Liu
}
\affiliations {
   Research Center for Social Computing and Information Retrieval, Harbin Institute of Technology, China\\
    \{hcwang, sdzhao\}@ir.hit.edu.cn
}

\usepackage{bibentry}

\begin{document}

\maketitle

\begin{abstract}

Large Language Models (LLMs) have demonstrated remarkable success in diverse natural language processing (NLP) tasks in general domains. However, LLMs sometimes generate responses with the \textit{hallucination} about medical facts due to limited domain knowledge. Such shortcomings pose potential risks in the utilization of LLMs within medical contexts. To address this challenge, we propose knowledge-tuning, which leverages structured medical knowledge bases for the LLMs to grasp domain knowledge efficiently and facilitate reliable response generation. We also release cMedKnowQA, a Chinese medical knowledge question-answering dataset constructed from medical knowledge bases to assess the medical knowledge proficiency of LLMs. Experimental results show that the LLMs which are knowledge-tuned with cMedKnowQA, can exhibit higher levels of accuracy in response generation compared with vanilla instruction-tuning and offer a new reliable way for the domain adaptation of LLMs.
\end{abstract}
\section{Introduction}
The advent of large language models (LLMs), representative by ChatGPT~\cite{chatgpt}, has generated significant interest due to their exceptional performance in understanding instructions and generating human-like responses. Compared to smaller models, LLMs exhibit strong generalization across various natural language processing (NLP) tasks and a unique emergent ability to solve unseen or complicated tasks. Despite ChatGPT's non-open source status, open-source communities have provided several alternatives, such as LLaMa~\cite{touvron2023llama}, with relatively affordable training costs. 

\begin{figure*}[ht] 
\centering 
\includegraphics[width=0.95\textwidth]{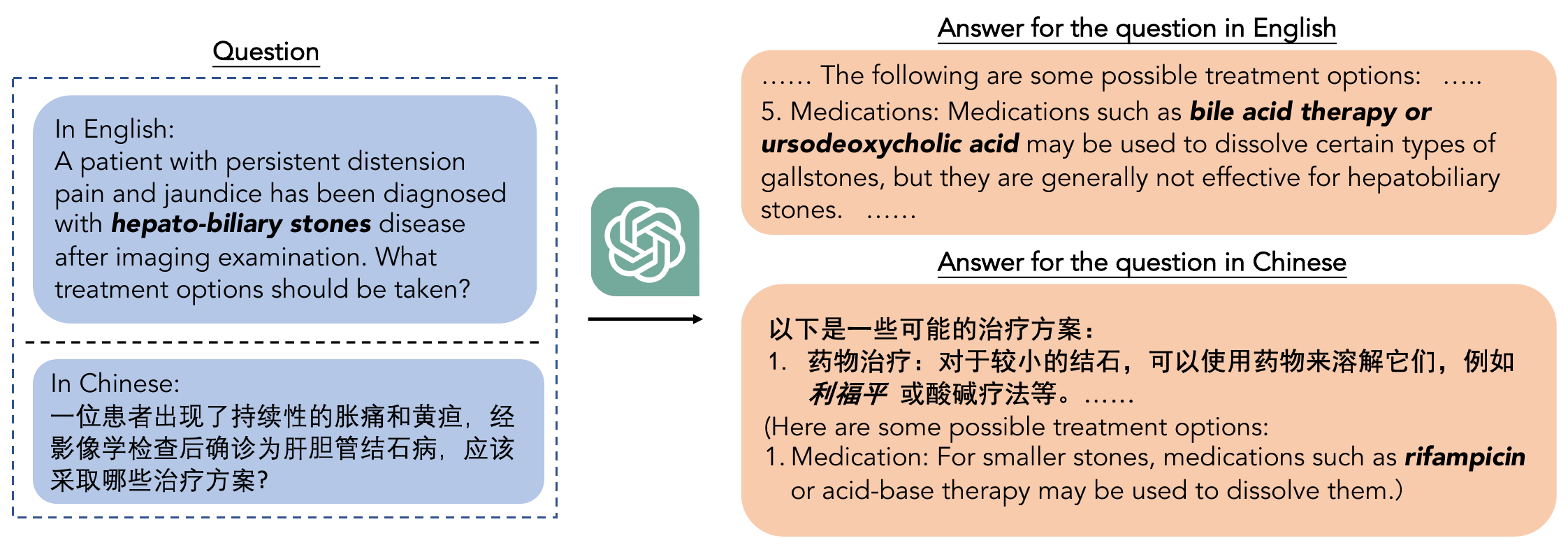} 
\caption{Response cases of ChatGPT with an \textit{identical} question raised in English and Chinese. ChatGPT provides incorrect medicine recommendations in response to the question in Chinese. Generated by ChatGPT on April 13th, 2023.} 
\label{intro_case} 
\end{figure*}

However, there is a dominant challenge for the adaption of the LLMs to the medical domain, that is the hallucination about the medical knowledge since LLMs are not designed to cater specifically to the medical domain. Their general domain knowledge often falls short when addressing such specialized fields, where accurate and domain-specific expert knowledge is critical, which leads to hallucination \cite{ji2023survey} in the model responses, especially for languages that are less well-trained than English. Figure \ref{intro_case} shows the responses generated by ChatGPT \cite{chatgpt} for an identical question in English and Chinese respectively. When answering the question in English, ChatGPT provides reasonable medications for ``hepatobiliary stones''. However, given the identical one in Chinese, ChatGPT recommends ``Rifampicin'', which is an antibiotic medicine to treat mycobacterial infections and not effective for hepatobiliary stones. Such hallucinations in the responses generated by the large language models can lead to sub-optimal drug recommendations and medical advice, potentially endangering patients.

Limited attempts have been undertaken to tackle the challenge, wherein current methodologies concentrate predominantly on equipping LLMs with medical data procured from real or synthetic dialogues \cite{yunxiang2023chatdoctor, xiong2023doctorglm} while the possibility of human fallacies is relatively high. Nevertheless, it is a nontrivial undertaking for LLMs to comprehend such knowledge only with the supervised fine-tuning approach and formulate replies that are consistent with knowledge and free from hallucinations. Recent studies made attempts to mitigate such problems by integrating the LLMs model with external API for specifically targeted fields \cite{shen2023hugginggpt,thoppilan2022lamda}. However, although there are various medical knowledge bases, such as UMLS \cite{mccray1995representation} and medical knowledge graphs \cite{Odmaa, li2020real}, there is no efficient API for structured medical knowledge that can be leveraged by LLMs till the present. 

In this study, we introduce ``knowledge-tuning'' that explicitly incorporates Chinese medical knowledge bases during both the training and inference phases of the LLMs. Knowledge-tuning first generates medical knowledge QA pairs based on structured knowledge bases through the ChatGPT API, and then trains the LLMs to generate the keyword and possible attributes as query parameters for each input. These query parameters are then used to retrieve relevant medical knowledge. Subsequently, the LLMs can generate responses for the input by referring to the retrieved medical knowledge. Consequently, the LLMs not only generate responses but also provide information about the knowledge source they rely on, thereby improving the quality and reliability of generated responses.

Our contributions can be summarized as follows: 

\begin{itemize}
    \item We introduce knowledge-tuning, an approach designed to effectively leverage structured medical domain knowledge into the responses generated by the LLMs, \textit{not only} mitigating the hallucination in the responses \textit{but also} providing the knowledge source, which is especially critical for the medical domain application.
    \item We develop an approach to constructing medical knowledge question-answer datasets with the knowledge bases and create the first Chinese medical knowledge question-answer dataset, cMedKnowQA.
    \item We propose a comprehensive evaluation metric from the aspects of accuracy of retrieved knowledge, helpfulness and harmlessness of the generated responses, to gauge the performance of the knowledge-tuning where traditional evaluation methods are not adequate enough. Experimental results demonstrate that knowledge-tuning shows a remarkable advantage compared with the baselines and remains effective in both the few-shot and generalization scenarios.
\end{itemize}

\section{Related Works}
\subsection{Large Language Models}
The considerable increase in language model scale has brought about substantial transformations in their quality, leading to the development of ChatGPT~\cite{chatgpt} and GPT-4.  has revolutionized the perception of LLMs by tackling the NLP tasks in a generation manner. These advancements have revolutionized the perception of LLMs by effectively addressing natural language processing (NLP) tasks in a generative manner. Despite the impressive performance exhibited by these models, OpenAI has not publicly disclosed specific details regarding their training methodologies or weight parameters. Consequently, several accessible LLMs, including LLaMA \cite{touvron2023llama}, Pythia \cite{biderman2023pythia}, and Bloom \cite{scao2022bloom}, have emerged as viable alternatives for research purposes. To enhance their performance, these models have employed techniques such as instruction-tuning \cite{wang2022self, alpaca, sanh2022multitask, chung2022scaling, weifinetuned} and reinforcement learning with human feedback \cite{ouyang2022training, bai2022training}, aiming to align the model outputs with human expectations. However, it should be noted that the instruction data primarily originate from iterate generations from LLMs, relying on only a few instruction seeds, which can introduce noise knowledge-related information.

\subsection{LLMs in Biomedical Domain}
Although LLMs exhibit remarkable performance in general domains, their lack of domain-specific knowledge results in sub-optimal performance in fields that require specialized expertise, such as bio-medicine. Several efforts have been made to adapt LLMs to the biomedical domain. Various trials have been made to pre-train the models with the corpora in the biomedical domain \cite {peng2019transfer,lee2020biobert,huang2019clinicalbert, luo2022biogpt}, learn domain-specific vocabulary for better representation \cite{lewis2020pretrained, gu2021domain}, or provide the LLMs with medical knowledge \cite{zhang2021smedbert, michalopoulos2021umlsbert,wang-etal-2022-prompt}. As for larger language models, base models have been instruct-tuned with synthesized biomedical conversations or real clinical dialogues \cite{yunxiang2023chatdoctor, xiong2023doctorglm, huatuogpt-2023}. These above works have illustrated the potential for LLMs to be successfully applied within the biomedical domain but the correctness of generated responses is only dependent on the embedded knowledge inside the LLMs.

\subsection{Tools for LLMs}
Since the ability of the LLMs can be limited in specific domains like mathematics and medicine \cite{thoppilan2022lamda}, various research has been dedicated to equipping LLMs with external tools. This methodology enables the delegation of tasks such as precise computation or information retrieval to external modules like a calculator or a search engine \cite{mialon2023augmented, thoppilan2022lamda}. Also, the integration of external sources enables the retrieval of natural language knowledge, as demonstrated by WebGPT \cite{nakano2021webgpt} and ReAct \cite{yao2022react} which utilize searching APIs. In addressing diverse NLP tasks, researchers turn to combining multi-source of APIs, models, plugins, and other tools \cite{schick2023toolformer,paranjape2023art, shen2023hugginggpt}. For the biomedical LLMs, the issue of response reliability holds significant importance and in this study, we investigate the medical knowledge function in facilitating reliable response generation for the LLMs.

\section{Methodology}
\begin{figure*}[ht] 
\centering 
\includegraphics[width=0.9\textwidth]{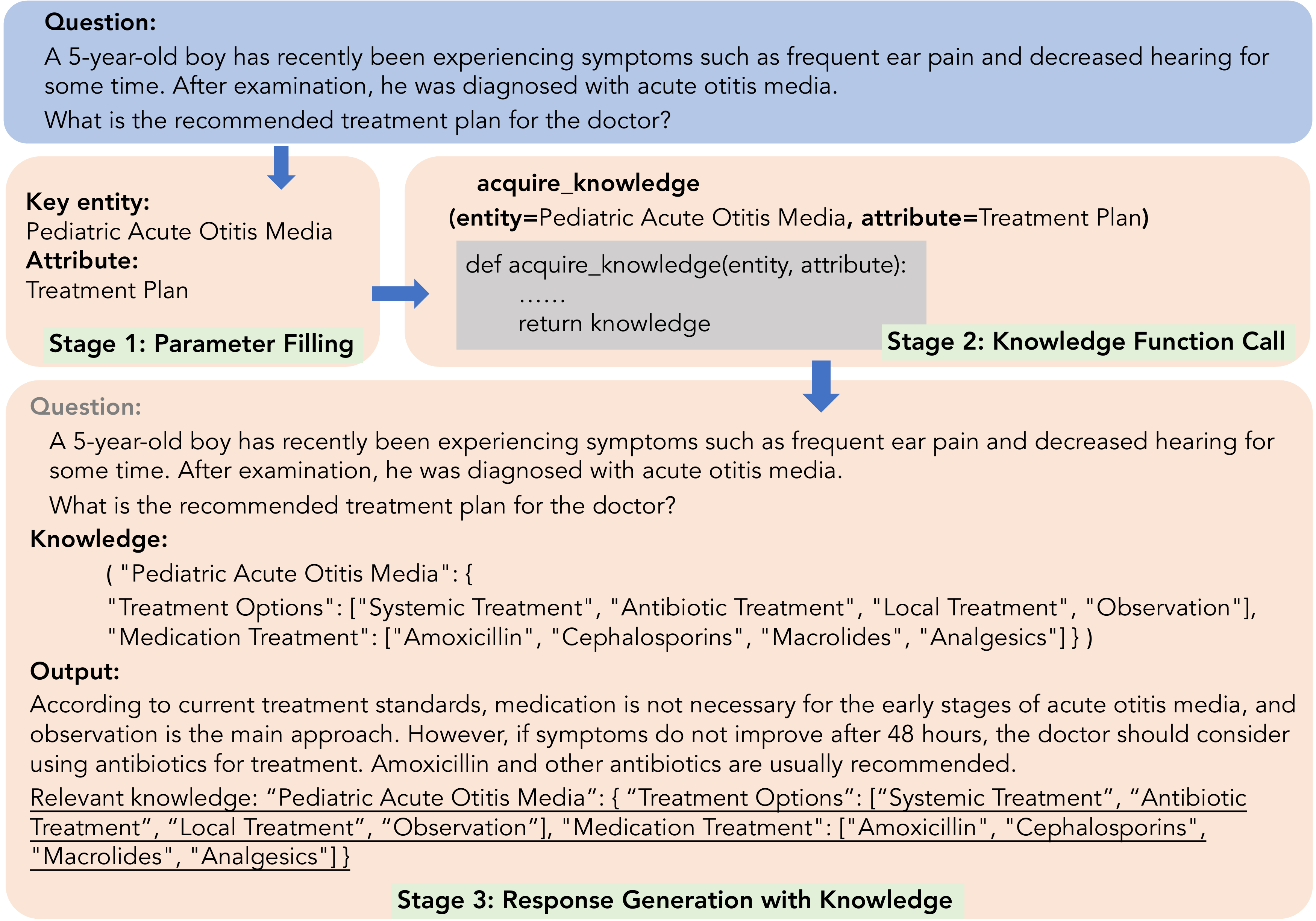} 
\caption{Process for knowledge-based response generation. Stage 1: Fill in the parameters for the knowledge retrieval based on the query question. Stage 2: Acquire the knowledge with filled parameters. Stage 3: Generate a response with acquired knowledge. Texts in Chinese have been translated into English.} 
\label{model_process_knwoweledge} 
\end{figure*}
Here, we first present the structured medical knowledge we utilize in this study. Subsequently, we transform the structured knowledge into training data suitable for the LLMs. Finally, we engage in knowledge-tuning that guides the LLMs to retrieve relevant medical knowledge in response to input queries and to generate responses based on the corresponding knowledge in a unified paradigm with LLMs during the training and inference stages, as illustrated in Figure \ref{model_process_knwoweledge}.

\begin{figure}[ht] 
\centering 
\includegraphics[width=0.9\columnwidth]{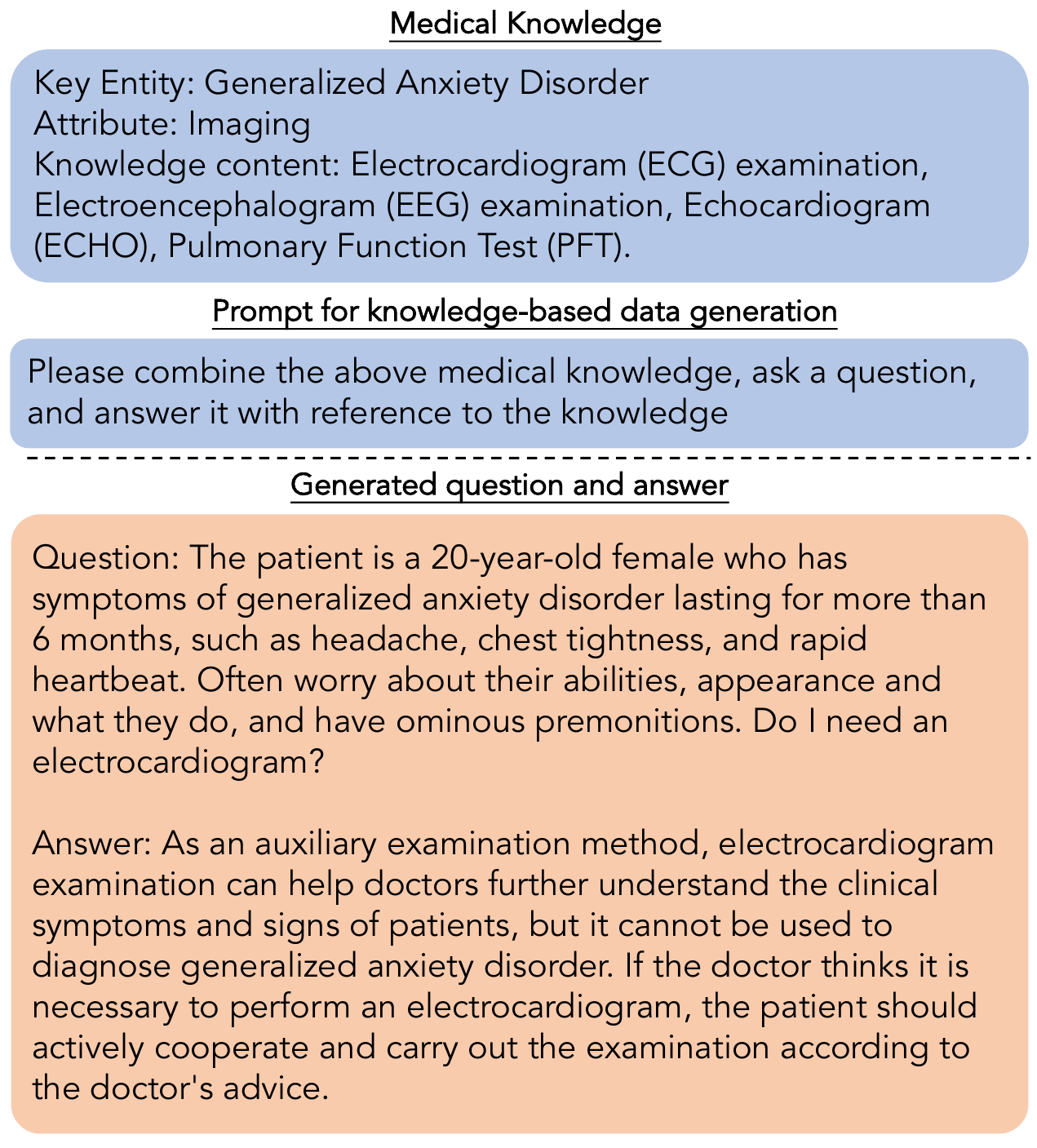} 
\caption{One medical knowledge-guided instance generated for knowledge-tuning. Texts in Chinese have been translated into English.} 
\label{instance} 
\end{figure}

\subsection{Structured Medical Knowledge Bases}
Medical knowledge generally includes structured medical knowledge like medical knowledge graphs or knowledge bases, and unstructured medical knowledge like medical guidelines and literature and in this study, we focus on the utilization of structured Chinese medical knowledge bases. Given a medical knowledge base $\mathcal{K}= \{k_1, k_2, \ldots, k_n \}$, each knowledge instance $k_i$ in the medical knowledge bases consists of a keyword of a medical entity $e_i$, the attribute of the entity $attr_i$ and the knowledge content $c_i$. Keywords contain ``diseases'', ``drugs'', ``symptoms'' and attributes contain ``pathogeny'', ``complication'', ``dosage'', etc. 

\begin{table*}[htbp]
\centering

\begin{tabularx}{0.98\textwidth}{p{0.09\textwidth}|p{0.282\textwidth}|p{0.518\textwidth}}
\toprule
\textbf{Source}    & \textbf{Knowledge in Chinese}   & \textbf{Knowledge translated to English} \\ \hline
CMeKG & \raisebox{\dimexpr-\height+\baselineskip}{%
  \includegraphics[width=0.28\textwidth]{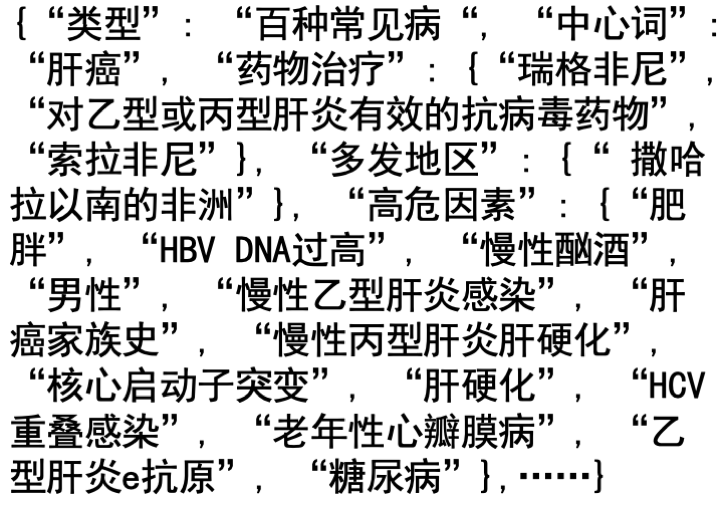}} & \{``class'': ``Common Diseases``, ``Key Word'': ``Liver Cancer``, ``Drug Treatment'': [``Regorafenib``, ``Antiviral drugs effective against hepatitis B or C``, ``Sorafenib``], ``High Prevalence Regions'': [``Sub-Saharan Africa``], ``High Risk Factors'': [``Obesity``, ``High HBV DNA levels``, ``Chronic alcoholism``, ``Male gender``, ``Chronic hepatitis B infection``, ``Family history of liver cancer``, ``Cirrhosis due to chronic hepatitis C``, ``Core promoter mutation``, ``Liver cirrhosis``, ``HCV co-infection``, ``Senile valvular heart disease``, ``Hepatitis B e antigen``, ``Diabetes``], ......\}.                                         \\ \hline
DingXiang Doctor    & \raisebox{\dimexpr-\height+\baselineskip}{%
  \includegraphics[width=0.28\textwidth]{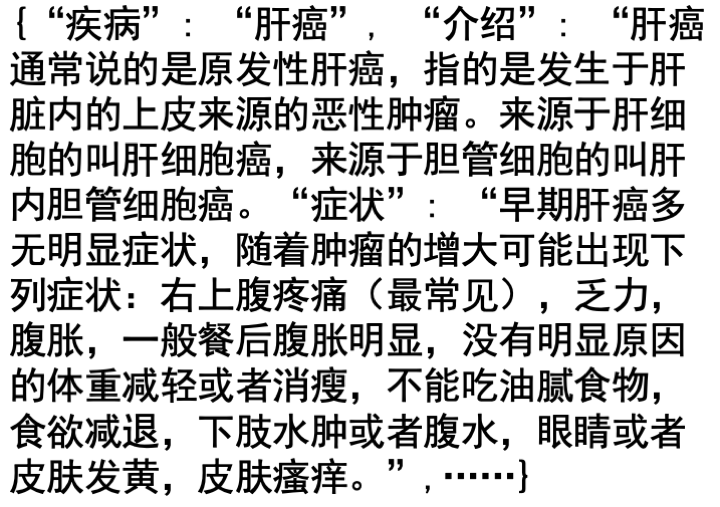}} &    \{``disease'': ``Liver Cancer``, ``introduction'': ``Liver cancer is a malignant tumor derived from the epithelial cells within the liver. Liver cell carcinoma refers to tumors derived from liver cells, ...... ``, ``symptoms'': ``In the early stages of liver cancer, there are often no obvious symptoms. As the tumor grows, the following symptoms may appear: pain in the upper right abdomen (most common), fatigue, bloating, significant bloating after meals, unexplained weight loss or thinning,......``,\}                                       \\ \bottomrule
\end{tabularx}
\caption{Knowledge cases in the CMeKG and DingXiang Doctor.}
\label{knowledge-label}
\end{table*}
\subsection{Knowledge-guided Data Generation}
\label{data-gen}
Instruction-tuning has been demonstrated to be a powerful technique for improving the performance of large language models \citep{weifinetuned, ouyang2022training}. Recently, some researchers \citep{wang2022self, alpaca} have proposed an alternative approach to generating instructions by utilizing language models, such as GPT-3 \citep{brown2020language} and ChatGPT \citep{chatgpt}, with the aid of a small number of seed instructions in the in-context Learning style. The models then generate the corresponding input query and output response pair $(q_i, r_i)$ guided with a provided instruction, which serves as training data for the language models.

However, in the biomedical domain, the outputs generated by LLMs are not always reliable, as illustrated in Figure \ref{intro_case} and it can be difficult to directly utilize the structured medical knowledge bases. Therefore, instead of generating training instances with instructions, we employ structured medical knowledge as guidance for creating knowledge-tuning data with LLMs. Given a piece of knowledge $k_i=(e_i, attr_i, c_i)$ in the medical knowledge base $\mathcal{K}$, we expect the ChatGPT API to produce a pair of model query and response $(q_i, r_i)$, which reflects the provided medical knowledge, using carefully crafted prompts. Within the medical domain, the model inputs are typically in the form of questions, while the expected model outputs usually consist of corresponding answers. Consequently, we fix the notion of ``instruction'' as a prompt template and retain solely the input and output components in our dataset, as demonstrated in Figure \ref{instance}.

In this way, each instance in the constructed dataset $\mathcal{D}$ is a tuple of $(e_i, attr_i, c_i, q_i, r_i)$. Although the quality of the data generated knowledge-guidingly surpasses that without guidance, the presence of noise data remains a concern. Thus, the constructed dataset undergoes initial assessment by ChatGPT itself and subsequent examination by experts in the field of medicine.

\subsection{Knowledge-tuning}
Given a knowledge-based dataset $\mathcal{D}$ where each instance is a tuple of $(e_i, attr_i, c_i, q_i, r_i)$, including a medical entity, entity attribute, the corresponding knowledge content and a query response pair relevant to the knowledge, \textit{knowledge-tuning} denotes training a base model to retrieve relevant knowledge of an input query relying on the generation of medical entity and attribute, and to generate corresponding response referring to the retrieved knowledge.

Initially, the medical entity $e_{pred}$ is predicted based on the input query $q$ with the entity prompt $\mathcal{P}_e$, expressed as 

\begin{equation}
\mathcal{M}(\mathcal{P}_e, q)=e_{pred}
\end{equation}

Subsequently, the attribute $attr_{pred}$ is generated by $\mathcal{M}$ using the input query $q$, the predicted entity $e_{pred}$ and the attribute prompt $\mathcal{P}_a$, denoted as 

\begin{equation}
\mathcal{M}(\mathcal{P}_a, q, e_{pred})=attr_{pred}
\end{equation}

Following this, the corresponding knowledge content $c$ is retrieved from the knowledge base using the parameter pair $(e_{pred}, attr_{pred})$. Finally, the LLMs generate a response $r$ given the input query $q$, the retrieved medical knowledge $c$ and the response prompt with knowledge $\mathcal{P}_{rk}$, as 

\begin{equation}
\mathcal{M}(\mathcal{P}_{rk}, q, c)=r_{pred}
\end{equation}

The loss function for knowledge-tuning $\mathcal{L}_{kt}$ comprises three components, which are losses of predicting $e_{pred}$, $attr_{pred}$ and $r_{pred}$, denoted as 
\begin{equation}
\mathcal{L}_{kt}=\mathcal{L}_{e}+\mathcal{L}_{attr}+\mathcal{L}_{r}
\end{equation}

The inference process adopts the same process above and is described in Algorithm \ref{alg:kt} in Appendix A. During the inference, the LLMs may encounter challenges in retrieving the relevant medical knowledge due to incorrect predictions of the entities or attributes. To address this issue, a dataset $\mathcal{D}^*$ is derived, where each instance contains $(q, r)$ and the LLMs are trained to respond solely with the input query, akin to instruction-tuning, denoted as 

\begin{equation}
\mathcal{M}(\mathcal{P}_{r}, q)=r_{pred}
\end{equation}
with the training loss of $\mathcal{L}_{it}$. Consequently, the overall loss function $\mathcal{L}$ is a combination of $\mathcal{L}_{kt}$ and $\mathcal{L}_{it}$, formulated as 
\begin{equation}
\mathcal{L} = \mathcal{L}_{kt} + \mathcal{L}_{it}
\end{equation}

\section{Experiment}
\subsection{Baselines and Implementations}
Knowledge-tuning, being a model-agnostic approach, is assessed utilizing two base models, namely Bloom \cite{scao2022bloom} and LLaMA \cite{touvron2023llama}. For our experiments, we specifically choose the LLaMA-based Alpaca-Chinese model \cite{chinese-llama-alpaca}, which involves expanding the vocabulary list and instruction-tuning using Chinese datasets, and align the Bloom models with Chinese instruction-tuning datasets. The base models comprising around 7 billion parameters are selected for a balance between performance and computational resources. 

For knowledge retrieval performance, knowledge-tuning is juxtaposed with two baselines: (1) BM25, a statistics-based  \cite{robertson2009probabilistic} and (2) Dense retrieval\cite{zhao2022dense}, wherein knowledge instances and input queries are vector-embedded, with retrieval based on maximum cosine similarity. Traditional NER techniques are infeasible due to potential query-entity mismatches.

To highlight the response generation performance of knowledge-tuning, we compare with the following baselines: (1) vanilla Bloom and Alpaca models; (2) instruction-tuning on both models using the $\mathcal{D}^*$ dataset; and (3) ChatGPT \cite{chatgpt}, an established OpenAI product. All models are optimized using LoRA \cite{hu2021lora}. See Appendix B for implementation specifics.

\subsection{Dataset}
We source the structured Chinese medical knowledge from two sources: (1) CMeKG, a Chinese medical knowledge base \cite{Odmaa}, which encompasses details about diseases, drugs, symptoms, among others, and (2) Chinese medical guidelines from DingXiang Doctor \footnote{\url{https://dxy.com}}. Table \ref{knowledge-label} illustrates cases from both CMeKG and guidelines of DingXiang Doctor. Using the ChatGPT API and these knowledge bases, we formulate a Chinese medical knowledge question-answer dataset, cMedKnowQA. Data curated with knowledge guidance outperforms that guided by instruction, though noise remains. To enhance data quality, medical experts have been employed to rigorously inspect and rectify inaccuracies. Consequently, cMedKnowQA features 7,449 entries, each presenting a question, answer, and relevant medical knowledge, divided into training, validation, and test sets at a 7:1:2 ratio.

\subsection{Metrics}
In the general domain, generative model assessment frequently employs metrics like Bleu and Rouge, gauging resemblance between model outputs and ground truths. Nevertheless, these metrics may not be apt for medical question-answering evaluations. Particularly in the biomedical realm, relying solely on output-ground truth similarity may not effectively capture the quality of generated answers. Figure \ref{intro_case} illustrates this limitation, where a model mistakenly suggests ``rifampicin'' for ``hepatobiliary stone''. Notably, even with such a significant error, metrics like Bleu still produce high scores, highlighting the limitations of similarity-centric evaluations in the biomedical field \cite{chang2023survey}.

Consequently, the efficacy of knowledge-tuning is critically examined from three distinct perspectives. (1) We introduce numeric metrics for the medical entity and knowledge as the accuracy of responses is often indicative of the entity and knowledge they reference. (2) Medical specialists assess the model outputs, offering a more nuanced evaluation than automatic metrics. (3) ChatGPT serves as an auxiliary evaluative standard. To ensure a thorough assessment of response quality, we advocate for the application of the $H_2$ (\textbf{H}elpfulness and \textbf{H}armlessness) score. The ``Helpfulness'' reflects the level of medical expertise exhibited in the model responses. Medical experts are tasked with rating the helpfulness of responses based on the relevant medical knowledge utilized, as opposed to relying on their own medical skills. This approach allows for a more accurate representation of how the LLMs leverage retrieved knowledge. Meanwhile, ``Harmlessness'' aims to identify any content within the responses that could potentially mislead users and put them in harm's way, such as erroneous medicine recommendations. An assortment of $H_2$ scoring illustrations by medical specialists can be located in Appendix C.

\subsection{Evaluation on the Medical Entity and Knowledge}
Table \ref{para_evaluation} illustrates the accuracy of the possible generated medical entity and knowledge by the LLMs. The baseline methods directly procure medical knowledge without the prediction of medical entities. Due to the vastness of the medical knowledge base and the intricacies of structured information, dense retrieval is limited to retrieve 2.6\% of all the knowledge. And BM25  manifests superior accuracy, approximately 55\%, in knowledge acquisition. 
Notably, the LLMs exhibit a noteworthy efficacy in the prediction of medical entities, with a recorded accuracy rate as high as 86.7\%. Pertaining to the retrieval of knowledge in conjunction with the forecasted entity and attribute, the LLMs demonstrate a precision rate reaching up to 71.4\%. This figure accounts for the occasional ambiguity in distinguishing specific candidate attributes.

\begin{table}[h]
\renewcommand{\arraystretch}{1.1}
\centering
\begin{tabular}{c|c|c}
\hline
\textbf{Base Model} & \begin{tabular}[c]{@{}c@{}}\textbf{Entity}\\ (Acc)\end{tabular} & \begin{tabular}[c]{@{}c@{}}\textbf{\textbf{Knowledge}}\\ (Acc)\end{tabular} \\ \hline
Random\dag & - & 0.013\\
\hline
Dense Retrieval & - & 2.6\\ \hline
BM25  & - & 54.9 \\
\hline

Alpaca  & 84.5   & 67.0        \\ 
\hline
Bloom  & 86.7 & 71.4 \\ 
\hline

\end{tabular}
\caption{Accuracy of generated medical entity and retrieved knowledge. \dag \ denotes the theoretical value.}
\label{para_evaluation}
\end{table}

\subsection{Evaluation on the Response Quality}
\paragraph{Medical Expert Evaluation}
The proposed metric, $H_2$, is employed to quantitatively evaluate the dimensions of ``Helpfulness'' and ``Harmlessness'' determined by medical professionals, each scaled between 1 and 3. Regarding the aspect of ``Helpfulness'', for the responses with retrieved medical knowledge, a rating of ``3'' delineates comprehensive coverage of the pertinent medical knowledge; a rating of ``2'' indicates the response, although omitting crucial information, remains effective; whereas a rating of ``1'' denotes a complete absence of helpfulness. For responses that do not encapsulate retrieved knowledge, they are graded between "3" and "1", symbolizing effectiveness, adequacy, and inacceptability in relation to the medical knowledge encompassed therein, respectively. In terms of ``Harmlessness'', a score of ``3'' signifies no harmful content in the response; ``2'' suggests the existence of erroneous yet non-detrimental information; and ``1'' underscores the inclusion of injurious information. Scoring principle of ``Harmlessness'' is similar to the above.

Two medical professionals are commissioned to appraise model-generated responses of 200 randomly selected queries from the test dataset. Utilizing Cohen's Kappa coefficient \cite{cohen1960coefficient}, an inter-rater agreement of 0.81 has been observed. The 
evaluations suggest that, while instruction-tuning emerges as a plausible method for adapting LLMs to the medical sphere \cite{wang2023huatuo, huatuogpt-2023}, it exhibits diminished efficacy when faced with intricate medical queries necessitating profound knowledge. Both the instruction-tuned and the original models exhibit subpar efficacy in the domains of helpfulness and harmlessness under such circumstances. Contrarily, knowledge-tuning augments the trustworthiness of the generated responses. A deeper exploration elucidates that knowledge-tuning, when paired with accurate knowledge retrieval, proffers superior outcomes, comparable to those of ChatGPT. A meticulous case study ensues in the subsequent section.

\paragraph{ChatGPT Evaluation}
We also incorporate ChatGPT to evaluate the identical subset employed in human assessments and instruct ChatGPT to categorize the responses into three distinct classifications: ``good'', ``moderate'' and ``bad'', numerically represented as ``3'', ``2'' and ``1'', respectively. As for the Bloom model, Table \ref{evaluation} shows that ChatGPT ranks the responses generated by instruction-tuned models as the lowest, with an average score of 2.61. The base model, on the other hand, manifests a marginally superior outcome with a score of 2.47, while the knowledge-tuned model achieves the highest score of 2.74. Such outcomes intimate that tuning the model with a knowledge QA dataset, especially with a restricted number of instances, might detrimentally influence its efficacy in confronting unfamiliar knowledge during the testing phase. Remarkably, upon solely analyzing the responses from knowledge-tuning that accurately retrieve  knowledge, the ChatGPT evaluation escalates to 2.79.

\begin{table}[h]
\renewcommand{\arraystretch}{1}
\centering
\begin{tabular}{l|c|c|c}
\hline
\multirow{2}{*}{\textbf{Base Model}} & \multicolumn{2}{c|}{$H_2$} &  \multirow{2}{*}{\begin{tabular}[c]{@{}c@{}}ChatGPT\\ Score $\uparrow$\end{tabular}} \\
\cline{2-3}
                            & $h_1 \uparrow$         & $h_2 \uparrow$        &  \\ \hline

Alpaca                          & 1.78   & 1.98       &   2.68 \\ 
Alpaca + \textit{i-tuning}      & 1.80   & 1.92      &  2.64    \\ 
Alpaca + \textit{k-tuning}      & 2.25   & 2.40      &   2.69   \\ 
Alpaca + \textit{k-tuning}\dag  & 2.63   & \textbf{2.59}   & \textbf{2.71}        \\ 
\hline
Bloom                           & 1.82   &  1.83     &  2.61   \\ 
Bloom + \textit{i-tuning}       & 1.81   & 1.88    &   2.47     \\ 
Bloom + \textit{k-tuning}       & 2.41   & 2.42      &  2.74    \\ 
Bloom + \textit{k-tuning}\dag   & \textbf{2.72}   & 2.57   &    \textbf{2.79}     \\ 
\hline
ChatGPT                         & 2.71   & \textbf{2.59}    & -     \\

\hline

\end{tabular}
\caption{Evaluation on the model responses. $H_2$ scores are the mean of two raters with a Cohen's kappa of 0.81. ``$h_1$'' indicates helpfulness and ``$h_2$'' denotes harmlessness. ``i-tuning'' denotes instruction-tuning. ``k-tuning'' denotes knowledge-tuning. \dag \ indicates knowledge-tuning with correct medical knowledge.}
\label{evaluation}
\end{table}

\section{Discussion}
\subsection{Case Study}
\begin{table*}[th]
\renewcommand{\arraystretch}{1}
\begin{tabularx}{0.95\textwidth}{X}
\toprule

\textbf{Medical Knowledge:} 
Entity: cicatricial pyloric obstruction, Attribute: symptom, Knowledge content: ``vomiting during afternoon and night, abdominal pain during the night and after eating.''                                                         \\ \hline
\textbf{Question:} 
When should we suspect a cicatricial pyloric obstruction?                                      
\\ \hline
\textbf{Golden response:} 
When patients have upper abdominal pain, postprandial vomiting, early satiety, and abdominal \textit{distension}, \textit{with or without weight loss}, one should suspect cicatricial pyloric obstruction.                                                                                                                                                                           \\ \hline
\textbf{Bloom:} 
When the following symptoms are present, suspect cicatricial pyloric obstruction: (1) Upper abdominal pain, (2) Nausea, (3) Vomiting, (4) Weight loss, (5) Belching, (6) \textit{\textbf{Black stools}}, (7) \textit{\textbf{Jaundice}}, ......                                                                                                                                                                   \\ \hline
\textbf{Bloom with instruction-tuning:} 
When the following symptoms are present, suspect scar-induced pyloric obstruction: abdominal pain, vomiting, \textit{acid reflux, \textbf{jaundice}, heartburn, belching, nausea, loss of appetite, weight loss, ......}.                                                                                                                             \\ \hline
\textbf{Bloom with knowledge-tuning:} 
(1) Abdominal pain and bloating: cicatricial pyloric obstruction often occurs after eating, especially at night. The upper abdominal pain worsens after meals and progresses to diffuse upper abdominal distension or discomfort as gastric retention occurs. (2) Vomiting: The most prominent symptom, vomiting mainly occurs in the afternoon and evening. \\ 
\bottomrule
\end{tabularx}
\caption{Case study for the Bloom model responses. Texts in Chinese have been translated into English. Golden response is generated by ChatGPT with the guidance of the medical knowledge. \textit{Italic} means contents with information not inside the provided medical knowledge. \textbf{\textit{Bold italic}} means wrong information.}
\label{case}
\end{table*}
Table \ref{case} presents a case study for the Bloom model with various tuning approaches. Given a medical knowledge concerning ``cicatricial pyloric obstruction'', ChatGPT is programmed to construct a paired question and golden response derived from this knowledge. However, it is observed that the generated responses by ChatGPT occasionally deviate from complete accuracy, as the model tends to incorporate additional information beyond the provided medical knowledge. Specifically, references to ``abdominal distension'' and ``weight loss'' are accurate yet exceed the confines of the provided knowledge, presenting potential pitfalls in response generation. Furthermore, the original Bloom model produces incorrect symptoms, such as ``Black stools'' and ``Jaundice''. Meanwhile, the Bloom model with instruction-tuning tends to enumerate symptoms pertinent to the digestive system, which are not inherently delineated in the given data. Conversely, the Bloom model with knowledge-tuning attempts to rephrase the medical knowledge into natural language while adding necessary embellishments, while still mostly adhering to the original knowledge.


\subsection{Few-shot Scenario for Knowledge-tuning}
\begin{figure}[ht] 
\centering 
\includegraphics[width=0.8\columnwidth]{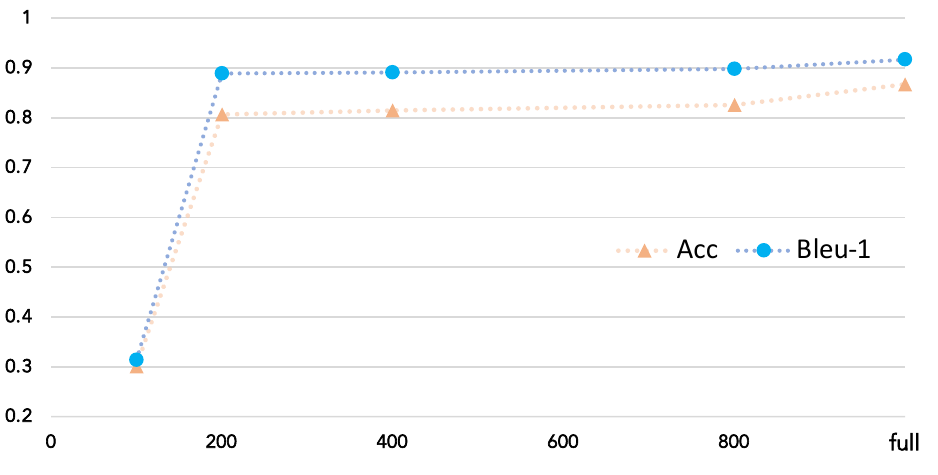} 
\caption{Entity generation in the few-shot scenarios.} 
\label{few-shot} 
\end{figure}

Since the efficacy of knowledge-tuning is inherently tied to the generation of medical entity and attribute, we investigate the utility of the Bloom model in the few-shot context for medical entity generation, using data subsets ranging from 100 to 800 instances from both training and validation sets. In addition to accuracy metrics, here we present the BLEU-1 score as a measure to gauge the caliber of entity generation as a reference. Figure \ref{few-shot} demonstrates the Bloom model underperforms at 100 instances. Yet, at 200 instances, its accuracy markedly improves to 80.7\%. When further refined using the complete datasets, the accuracy escalates to 86.7\%. This underscores the potential of knowledge-tuning in addressing few-shot scenarios, particularly when datasets may inadequately represent newly emerging or rare diseases.

\subsection{Generalization with Unseen Entities}
\begin{figure}[ht] 
\centering 
\includegraphics[width=0.8\columnwidth]{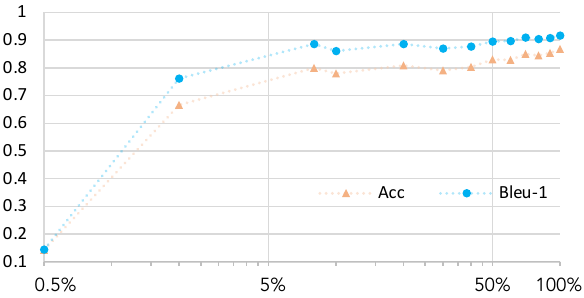} 
\caption{Model generalization with unseen entities. X-axis indicates the partition of seen entities in the training set.} 
\label{gen} 
\end{figure}

As the cMedKnowQA dataset contains medical entities that are associated with multiple attributes, we conduct experiments wherein only a portion of the medical entities presented in the test set have been seen during training, to assess the model generalization capability on unseen entities. In order to achieve this, we utilize the entire dataset as the test set and organize it based on medical entities. Subsequently, we create training sets by sampling distinct medical entities from 0.05\% to 60\%. The test set remains constant throughout the process. The results, as illustrated in Figure \ref{gen}, indicate that the Bloom model exhibits inadequate performance when trained on extremely limited data. However, it demonstrates robust generalization capabilities when trained with no less than 0.05\% of distinct entities. This suggests that the knowledge-tuned model can facilitate knowledge transfer to new medical entities without requiring further tuning.

\section{Conclusion}
In this paper, we address the issue of knowledge inaccuracy of medical facts in responses generated by LLMs which is critical for the application in the medical domain, particularly in the context of Chinese. We propose a novel approach called knowledge-tuning, which utilizes a medical knowledge function as a plug-in helper for LLMs to efficiently grasp domain knowledge and enhance the reliability of response generation. With experiments on our proposed Chinese medical knowledge QA dataset, cMedKnowQA, our knowledge-tuned model achieves higher accuracy and reliability in generating responses and also shows consistency with less training data and generalization with unseen entities thereby enlightening the domain adaptation of LLMs.

\clearpage
\newpage

\bibliography{aaai24,custom}

\begin{thebibliography}{42}
\providecommand{\natexlab}[1]{#1}

\bibitem[{Bai et~al.(2022)Bai, Jones, Ndousse, Askell, Chen, DasSarma, Drain,
  Fort, Ganguli, Henighan et~al.}]{bai2022training}
Bai, Y.; Jones, A.; Ndousse, K.; Askell, A.; Chen, A.; DasSarma, N.; Drain, D.;
  Fort, S.; Ganguli, D.; Henighan, T.; et~al. 2022.
\newblock Training a helpful and harmless assistant with reinforcement learning
  from human feedback.
\newblock \emph{arXiv preprint arXiv:2204.05862}.

\bibitem[{Biderman et~al.(2023)Biderman, Schoelkopf, Anthony, Bradley, O'Brien,
  Hallahan, Khan, Purohit, Prashanth, Raff, Skowron, Sutawika, and van~der
  Wal}]{biderman2023pythia}
Biderman, S.; Schoelkopf, H.; Anthony, Q.; Bradley, H.; O'Brien, K.; Hallahan,
  E.; Khan, M.~A.; Purohit, S.; Prashanth, U.~S.; Raff, E.; Skowron, A.;
  Sutawika, L.; and van~der Wal, O. 2023.
\newblock Pythia: A Suite for Analyzing Large Language Models Across Training
  and Scaling.
\newblock arXiv:2304.01373.

\bibitem[{Brown et~al.(2020)Brown, Mann, Ryder, Subbiah, Kaplan, Dhariwal,
  Neelakantan, Shyam, Sastry, Askell et~al.}]{brown2020language}
Brown, T.; Mann, B.; Ryder, N.; Subbiah, M.; Kaplan, J.~D.; Dhariwal, P.;
  Neelakantan, A.; Shyam, P.; Sastry, G.; Askell, A.; et~al. 2020.
\newblock Language models are few-shot learners.
\newblock \emph{Advances in neural information processing systems}, 33:
  1877--1901.

\bibitem[{Chang et~al.(2023)Chang, Wang, Wang, Wu, Zhu, Chen, Yang, Yi, Wang,
  Wang et~al.}]{chang2023survey}
Chang, Y.; Wang, X.; Wang, J.; Wu, Y.; Zhu, K.; Chen, H.; Yang, L.; Yi, X.;
  Wang, C.; Wang, Y.; et~al. 2023.
\newblock A Survey on Evaluation of Large Language Models.
\newblock \emph{arXiv preprint arXiv:2307.03109}.

\bibitem[{Chung et~al.(2022)Chung, Hou, Longpre, Zoph, Tay, Fedus, Li, Wang,
  Dehghani, Brahma et~al.}]{chung2022scaling}
Chung, H.~W.; Hou, L.; Longpre, S.; Zoph, B.; Tay, Y.; Fedus, W.; Li, E.; Wang,
  X.; Dehghani, M.; Brahma, S.; et~al. 2022.
\newblock Scaling instruction-finetuned language models.
\newblock \emph{arXiv preprint arXiv:2210.11416}.

\bibitem[{Cohen(1960)}]{cohen1960coefficient}
Cohen, J. 1960.
\newblock A coefficient of agreement for nominal scales.
\newblock \emph{Educational and psychological measurement}, 20(1): 37--46.

\bibitem[{Cui, Yang, and Yao(2023)}]{chinese-llama-alpaca}
Cui, Y.; Yang, Z.; and Yao, X. 2023.
\newblock Efficient and Effective Text Encoding for Chinese LLaMA and Alpaca.
\newblock \emph{arXiv preprint arXiv:2304.08177}.

\bibitem[{Gu et~al.(2021)Gu, Tinn, Cheng, Lucas, Usuyama, Liu, Naumann, Gao,
  and Poon}]{gu2021domain}
Gu, Y.; Tinn, R.; Cheng, H.; Lucas, M.; Usuyama, N.; Liu, X.; Naumann, T.; Gao,
  J.; and Poon, H. 2021.
\newblock Domain-specific language model pretraining for biomedical natural
  language processing.
\newblock \emph{ACM Transactions on Computing for Healthcare (HEALTH)}, 3(1):
  1--23.

\bibitem[{Hu et~al.(2021)Hu, Wallis, Allen-Zhu, Li, Wang, Wang, Chen
  et~al.}]{hu2021lora}
Hu, E.~J.; Wallis, P.; Allen-Zhu, Z.; Li, Y.; Wang, S.; Wang, L.; Chen, W.;
  et~al. 2021.
\newblock LoRA: Low-Rank Adaptation of Large Language Models.
\newblock In \emph{International Conference on Learning Representations}.

\bibitem[{Huang, Altosaar, and Ranganath(2019)}]{huang2019clinicalbert}
Huang, K.; Altosaar, J.; and Ranganath, R. 2019.
\newblock Clinicalbert: Modeling clinical notes and predicting hospital
  readmission.
\newblock \emph{arXiv preprint arXiv:1904.05342}.

\bibitem[{Ji et~al.(2023)Ji, Lee, Frieske, Yu, Su, Xu, Ishii, Bang, Madotto,
  and Fung}]{ji2023survey}
Ji, Z.; Lee, N.; Frieske, R.; Yu, T.; Su, D.; Xu, Y.; Ishii, E.; Bang, Y.~J.;
  Madotto, A.; and Fung, P. 2023.
\newblock Survey of hallucination in natural language generation.
\newblock \emph{ACM Computing Surveys}, 55(12): 1--38.

\bibitem[{Lee et~al.(2020)Lee, Yoon, Kim, Kim, Kim, So, and
  Kang}]{lee2020biobert}
Lee, J.; Yoon, W.; Kim, S.; Kim, D.; Kim, S.; So, C.~H.; and Kang, J. 2020.
\newblock BioBERT: a pre-trained biomedical language representation model for
  biomedical text mining.
\newblock \emph{Bioinformatics}, 36(4): 1234--1240.

\bibitem[{Lewis et~al.(2020)Lewis, Ott, Du, and Stoyanov}]{lewis2020pretrained}
Lewis, P.; Ott, M.; Du, J.; and Stoyanov, V. 2020.
\newblock Pretrained Language Models for Biomedical and Clinical Tasks:
  Understanding and Extending the State-of-the-Art.
\newblock In \emph{Proceedings of the 3rd Clinical Natural Language Processing
  Workshop}, 146--157.

\bibitem[{Li et~al.(2020)Li, Wang, Yan, Wang, Li, Jiang, Sun, Tang, Chang, Wang
  et~al.}]{li2020real}
Li, L.; Wang, P.; Yan, J.; Wang, Y.; Li, S.; Jiang, J.; Sun, Z.; Tang, B.;
  Chang, T.-H.; Wang, S.; et~al. 2020.
\newblock Real-world data medical knowledge graph: construction and
  applications.
\newblock \emph{Artificial intelligence in medicine}, 103: 101817.

\bibitem[{Li et~al.(2023)Li, Li, Zhang, Dan, and
  Zhang}]{yunxiang2023chatdoctor}
Li, Y.; Li, Z.; Zhang, K.; Dan, R.; and Zhang, Y. 2023.
\newblock ChatDoctor: A Medical Chat Model Fine-tuned on LLaMA Model using
  Medical Domain Knowledge.
\newblock arXiv:2303.14070.

\bibitem[{Luo et~al.(2022)Luo, Sun, Xia, Qin, Zhang, Poon, and
  Liu}]{luo2022biogpt}
Luo, R.; Sun, L.; Xia, Y.; Qin, T.; Zhang, S.; Poon, H.; and Liu, T.-Y. 2022.
\newblock BioGPT: generative pre-trained transformer for biomedical text
  generation and mining.
\newblock \emph{Briefings in Bioinformatics}, 23(6): bbac409.

\bibitem[{McCray and Nelson(1995)}]{mccray1995representation}
McCray, A.~T.; and Nelson, S.~J. 1995.
\newblock The representation of meaning in the UMLS.
\newblock \emph{Methods of information in medicine}, 34(01/02): 193--201.

\bibitem[{Mialon et~al.(2023)Mialon, Dess{\`\i}, Lomeli, Nalmpantis, Pasunuru,
  Raileanu, Rozi{\`e}re, Schick, Dwivedi-Yu, Celikyilmaz
  et~al.}]{mialon2023augmented}
Mialon, G.; Dess{\`\i}, R.; Lomeli, M.; Nalmpantis, C.; Pasunuru, R.; Raileanu,
  R.; Rozi{\`e}re, B.; Schick, T.; Dwivedi-Yu, J.; Celikyilmaz, A.; et~al.
  2023.
\newblock Augmented language models: a survey.
\newblock \emph{arXiv preprint arXiv:2302.07842}.

\bibitem[{Michalopoulos et~al.(2021)Michalopoulos, Wang, Kaka, Chen, and
  Wong}]{michalopoulos2021umlsbert}
Michalopoulos, G.; Wang, Y.; Kaka, H.; Chen, H.; and Wong, A. 2021.
\newblock UmlsBERT: Clinical Domain Knowledge Augmentation of Contextual
  Embeddings Using the Unified Medical Language System Metathesaurus.
\newblock In \emph{Proceedings of the 2021 Conference of the North American
  Chapter of the Association for Computational Linguistics: Human Language
  Technologies}, 1744--1753.

\bibitem[{Nakano et~al.(2021)Nakano, Hilton, Balaji, Wu, Ouyang, Kim, Hesse,
  Jain, Kosaraju, Saunders et~al.}]{nakano2021webgpt}
Nakano, R.; Hilton, J.; Balaji, S.; Wu, J.; Ouyang, L.; Kim, C.; Hesse, C.;
  Jain, S.; Kosaraju, V.; Saunders, W.; et~al. 2021.
\newblock Webgpt: Browser-assisted question-answering with human feedback.
\newblock \emph{arXiv preprint arXiv:2112.09332}.

\bibitem[{Odmaa et~al.(2019)Odmaa, Yunfei, Zhifang, Damai, Baobao, Sujian, and
  Hongying}]{Odmaa}
Odmaa, B.; Yunfei, Y.; Zhifang, S.; Damai, D.; Baobao, C.; Sujian, L.; and
  Hongying, Z. 2019.
\newblock Preliminary study on the construction of Chinese medical knowledge
  graph.
\newblock \emph{Journal of Chinese Information Processing}, 33(10): 1--7.

\bibitem[{OpenAI(2022)}]{chatgpt}
OpenAI. 2022.
\newblock ChatGPT.
\newblock \url{https://chat.openai.com}.

\bibitem[{Ouyang et~al.(2022)Ouyang, Wu, Jiang, Almeida, Wainwright, Mishkin,
  Zhang, Agarwal, Slama, Ray et~al.}]{ouyang2022training}
Ouyang, L.; Wu, J.; Jiang, X.; Almeida, D.; Wainwright, C.; Mishkin, P.; Zhang,
  C.; Agarwal, S.; Slama, K.; Ray, A.; et~al. 2022.
\newblock Training language models to follow instructions with human feedback.
\newblock \emph{Advances in Neural Information Processing Systems}, 35:
  27730--27744.

\bibitem[{Paranjape et~al.(2023)Paranjape, Lundberg, Singh, Hajishirzi,
  Zettlemoyer, and Ribeiro}]{paranjape2023art}
Paranjape, B.; Lundberg, S.; Singh, S.; Hajishirzi, H.; Zettlemoyer, L.; and
  Ribeiro, M.~T. 2023.
\newblock ART: Automatic multi-step reasoning and tool-use for large language
  models.
\newblock \emph{arXiv preprint arXiv:2303.09014}.

\bibitem[{Peng, Yan, and Lu(2019)}]{peng2019transfer}
Peng, Y.; Yan, S.; and Lu, Z. 2019.
\newblock Transfer Learning in Biomedical Natural Language Processing: An
  Evaluation of BERT and ELMo on Ten Benchmarking Datasets.
\newblock In \emph{Proceedings of the 18th BioNLP Workshop and Shared Task},
  58--65.

\bibitem[{Robertson, Zaragoza et~al.(2009)}]{robertson2009probabilistic}
Robertson, S.; Zaragoza, H.; et~al. 2009.
\newblock The probabilistic relevance framework: BM25 and beyond.
\newblock \emph{Foundations and Trends{\textregistered} in Information
  Retrieval}, 3(4): 333--389.

\bibitem[{Sanh et~al.(2022)Sanh, Webson, Raffel, Bach, Sutawika, Alyafeai,
  Chaffin, Stiegler, Le~Scao, Raja et~al.}]{sanh2022multitask}
Sanh, V.; Webson, A.; Raffel, C.; Bach, S.~H.; Sutawika, L.; Alyafeai, Z.;
  Chaffin, A.; Stiegler, A.; Le~Scao, T.; Raja, A.; et~al. 2022.
\newblock Multitask Prompted Training Enables Zero-Shot Task Generalization.
\newblock In \emph{Tenth International Conference on Learning Representations
  (ICLR)}.

\bibitem[{Scao et~al.(2022)Scao, Fan, Akiki, Pavlick, Ili{\'c}, Hesslow,
  Castagn{\'e}, Luccioni, Yvon, Gall{\'e} et~al.}]{scao2022bloom}
Scao, T.~L.; Fan, A.; Akiki, C.; Pavlick, E.; Ili{\'c}, S.; Hesslow, D.;
  Castagn{\'e}, R.; Luccioni, A.~S.; Yvon, F.; Gall{\'e}, M.; et~al. 2022.
\newblock Bloom: A 176b-parameter open-access multilingual language model.
\newblock \emph{arXiv preprint arXiv:2211.05100}.

\bibitem[{Schick et~al.(2023)Schick, Dwivedi-Yu, Dess{\`\i}, Raileanu, Lomeli,
  Zettlemoyer, Cancedda, and Scialom}]{schick2023toolformer}
Schick, T.; Dwivedi-Yu, J.; Dess{\`\i}, R.; Raileanu, R.; Lomeli, M.;
  Zettlemoyer, L.; Cancedda, N.; and Scialom, T. 2023.
\newblock Toolformer: Language models can teach themselves to use tools.
\newblock \emph{arXiv preprint arXiv:2302.04761}.

\bibitem[{Shen et~al.(2023)Shen, Song, Tan, Li, Lu, and
  Zhuang}]{shen2023hugginggpt}
Shen, Y.; Song, K.; Tan, X.; Li, D.; Lu, W.; and Zhuang, Y. 2023.
\newblock Hugginggpt: Solving ai tasks with chatgpt and its friends in
  huggingface.
\newblock \emph{arXiv preprint arXiv:2303.17580}.

\bibitem[{Taori et~al.(2023)Taori, Gulrajani, Zhang, Dubois, Li, Guestrin,
  Liang, and Hashimoto}]{alpaca}
Taori, R.; Gulrajani, I.; Zhang, T.; Dubois, Y.; Li, X.; Guestrin, C.; Liang,
  P.; and Hashimoto, T.~B. 2023.
\newblock Stanford Alpaca: An Instruction-following LLaMA model.
\newblock \url{https://github.com/tatsu-lab/stanford_alpaca}.

\bibitem[{Thoppilan et~al.(2022)Thoppilan, De~Freitas, Hall, Shazeer,
  Kulshreshtha, Cheng, Jin, Bos, Baker, Du et~al.}]{thoppilan2022lamda}
Thoppilan, R.; De~Freitas, D.; Hall, J.; Shazeer, N.; Kulshreshtha, A.; Cheng,
  H.-T.; Jin, A.; Bos, T.; Baker, L.; Du, Y.; et~al. 2022.
\newblock Lamda: Language models for dialog applications.
\newblock \emph{arXiv preprint arXiv:2201.08239}.

\bibitem[{Touvron et~al.(2023)Touvron, Lavril, Izacard, Martinet, Lachaux,
  Lacroix, Rozi{\`e}re, Goyal, Hambro, Azhar et~al.}]{touvron2023llama}
Touvron, H.; Lavril, T.; Izacard, G.; Martinet, X.; Lachaux, M.-A.; Lacroix,
  T.; Rozi{\`e}re, B.; Goyal, N.; Hambro, E.; Azhar, F.; et~al. 2023.
\newblock Llama: Open and efficient foundation language models.
\newblock \emph{arXiv preprint arXiv:2302.13971}.

\bibitem[{Wang et~al.(2023)Wang, Liu, Xi, Qiang, Zhao, Qin, and
  Liu}]{wang2023huatuo}
Wang, H.; Liu, C.; Xi, N.; Qiang, Z.; Zhao, S.; Qin, B.; and Liu, T. 2023.
\newblock HuaTuo: Tuning LLaMA Model with Chinese Medical Knowledge.
\newblock \emph{arXiv preprint arXiv:2304.06975}.

\bibitem[{Wang et~al.(2022{\natexlab{a}})Wang, Liu, Xi, Zhao, Ju, Zhang, Zhang,
  Zheng, Qin, and Liu}]{wang-etal-2022-prompt}
Wang, H.; Liu, C.; Xi, N.; Zhao, S.; Ju, M.; Zhang, S.; Zhang, Z.; Zheng, Y.;
  Qin, B.; and Liu, T. 2022{\natexlab{a}}.
\newblock Prompt Combines Paraphrase: Teaching Pre-trained Models to Understand
  Rare Biomedical Words.
\newblock In \emph{Proceedings of the 29th International Conference on
  Computational Linguistics}, 1422--1431. International Committee on
  Computational Linguistics.

\bibitem[{Wang et~al.(2022{\natexlab{b}})Wang, Kordi, Mishra, Liu, Smith,
  Khashabi, and Hajishirzi}]{wang2022self}
Wang, Y.; Kordi, Y.; Mishra, S.; Liu, A.; Smith, N.~A.; Khashabi, D.; and
  Hajishirzi, H. 2022{\natexlab{b}}.
\newblock Self-Instruct: Aligning Language Model with Self Generated
  Instructions.
\newblock \emph{arXiv preprint arXiv:2212.10560}.

\bibitem[{Wei et~al.(2022)Wei, Bosma, Zhao, Guu, Yu, Lester, Du, Dai, and
  Le}]{weifinetuned}
Wei, J.; Bosma, M.; Zhao, V.; Guu, K.; Yu, A.~W.; Lester, B.; Du, N.; Dai,
  A.~M.; and Le, Q.~V. 2022.
\newblock Finetuned Language Models are Zero-Shot Learners.
\newblock In \emph{International Conference on Learning Representations}.

\bibitem[{Xiong et~al.(2023)Xiong, Wang, Zhu, Zhao, Liu, Wang, and
  Shen}]{xiong2023doctorglm}
Xiong, H.; Wang, S.; Zhu, Y.; Zhao, Z.; Liu, Y.; Wang, Q.; and Shen, D. 2023.
\newblock Doctorglm: Fine-tuning your chinese doctor is not a herculean task.
\newblock \emph{arXiv preprint arXiv:2304.01097}.

\bibitem[{Yao et~al.(2022)Yao, Zhao, Yu, Du, Shafran, Narasimhan, and
  Cao}]{yao2022react}
Yao, S.; Zhao, J.; Yu, D.; Du, N.; Shafran, I.; Narasimhan, K.; and Cao, Y.
  2022.
\newblock React: Synergizing reasoning and acting in language models.
\newblock \emph{arXiv preprint arXiv:2210.03629}.

\bibitem[{Zhang et~al.(2023)Zhang, Chen, Jiang, Yu, Chen, Li, Chen, Wu, Zhang,
  Xiao, Wan, Wang, and Li}]{huatuogpt-2023}
Zhang, H.; Chen, J.; Jiang, F.; Yu, F.; Chen, Z.; Li, J.; Chen, G.; Wu, X.;
  Zhang, Z.; Xiao, Q.; Wan, X.; Wang, B.; and Li, H. 2023.
\newblock HuatuoGPT, Towards Taming Language Models To Be a Doctor.
\newblock \emph{arXiv preprint arXiv:2305.15075}.

\bibitem[{Zhang et~al.(2021)Zhang, Cai, Wang, Qiu, Yang, and
  He}]{zhang2021smedbert}
Zhang, T.; Cai, Z.; Wang, C.; Qiu, M.; Yang, B.; and He, X. 2021.
\newblock SMedBERT: A Knowledge-Enhanced Pre-trained Language Model with
  Structured Semantics for Medical Text Mining.
\newblock In \emph{Proceedings of the 59th Annual Meeting of the Association
  for Computational Linguistics and the 11th International Joint Conference on
  Natural Language Processing (Volume 1: Long Papers)}, 5882--5893.

\bibitem[{Zhao et~al.(2022)Zhao, Liu, Ren, and Wen}]{zhao2022dense}
Zhao, W.~X.; Liu, J.; Ren, R.; and Wen, J.-R. 2022.
\newblock Dense text retrieval based on pretrained language models: A survey.
\newblock \emph{arXiv preprint arXiv:2211.14876}.

\end{thebibliography}
\clearpage
\newpage
\appendix
\section{A. Knowledge-tuning Algorithm}
\begin{algorithm}[hbtp]
\begin{small}
\caption{\textsc{Knowledge-tuning} during inference}
\label{alg:kt}
\begin{algorithmic}[1]

\Require $\mathcal{D}=\{k_i=(e_i, attr_i, c_i, q_i, r_i)\}, i=1,2,...,n$
\Ensure $\mathcal{M}$, \textbf{Test set:} $\mathcal{T}$

\Function{Knowledge}{$\mathcal{D}, e_{pred}, attr_{pred}$}
    \If{$(e_{pred}, attr_{pred}) \in \mathcal{D}|_{e,attr}$}
        \State \textbf{return} $c$
    \Else
        \State \textbf{return} 
    \EndIf
\EndFunction

\Function{Inference}{$\mathcal{M}, \mathcal{T}$}
    \For{$q_{test} \in \mathcal{T}$)}
        \State $e_{test} \gets \mathcal{M}(\mathcal{P}_e, q_{test})$
        \State $attr_{test} \gets \mathcal{M}(\mathcal{P}_a, q_{test}, e_{test})$
        \State $c_{test} \gets$ \Call{Knowledge}{$e_{test}, attr_{test}$}
        \If{$c_{test}$} 
            \State $r_{test} \gets (\mathcal{M}(\mathcal{P}_{rk}, q_{test}, c_{test}), c_{test})$
        \Else
            \State $r_{test} \gets \mathcal{M}(\mathcal{P}_r, q_{test})$
        \EndIf
    \EndFor
\EndFunction

\end{algorithmic}\end{small}
\end{algorithm}

\section{B. Implementations}
As for the knowledge-tuning and instruction-tuning applied to the Alpaca and Bloom models, we employ LoRA to achieve parameter-efficiency tuning. For the main experiment, the hyperparameters are configured as presented in Table \ref{hyper}. All experiments are conducted utilizing NVIDIA A100-80GB-PCIEs. 

Prompt templates involved in this study are as follows (texts in Chinese have been translated into English): (1) Prompt for the ChatGPT API to generate knowledge-based question-answer pairs: ``Please combine the above medical knowledge, ask a question, and answer it with reference to the knowledge''. (2) Prompt $\mathcal{P}_e$ for the entity generation: ``What medical entity, like disease or drug, is mentioned in this question?'' (3) Prompt $\mathcal{P}_{attr}$ for the entity generation: ``What attribute about the {medical entity} is mentioned in this question?'' (4) Prompt $\mathcal{P}_{rk}$ for the knowledge-guided response generation: ``Answer the question with the above medical knowledge''. (5) Prompt $\mathcal{P}_r$ for the response generation: ``Answer the question with the above medical knowledge''.

\begin{table}[htbp]
\centering
\begin{tabular}{l|l}
\hline
\textbf{Hyperparameters}     & \textbf{Values}           \\
\hline
Batch size          & 128              \\
Max epoch           & 10               \\
Learning rate       & 5e-4             \\
LoRA rank           & 8                \\
LoRA alpha          & 16               \\
LoRA dropout        & 0.05             \\
LoRA target modules & q\_proj, v\_proj \\
\hline
\end{tabular}
\caption{Hyperparameters in the main experiments.}
\label{hyper}
\end{table}

\section{C. Medical Specialist $H_2$ Scoring Cases}
We display several $H_2$ score cases by the medical specialists in Table \ref{h2}.
\begin{table*}[ht]
\centering
\renewcommand{\arraystretch}{1.06}
\begin{tabularx}{\textwidth}{p{0.88\textwidth}|p{0.025\textwidth}|p{0.02\textwidth}}

\toprule
 & \multicolumn{2}{c}{$H_2$} \\ \cline{2-3}
 & $h_1$         & $h_2$ \\ \hline
\textbf{Medical Knowledge:} 
Entity: acetaminophen dihydrocodeine, Attribute: adverse reactions, Knowledge content: ``Agitated, nausea, vomiting, constipation, dizziness, seizure, excitement.''                  & -         & -                               \\ \hline
\textbf{Question:} 
What are the common adverse reactions to acetaminophen dihydrocodeine?                                      
 & - & -   \\ \hline

\textbf{Bloom:} 
The adverse reactions of acetaminophen and dihydrocodeine are mainly as follows:
Gastrointestinal reactions: Nausea, vomiting, abdominal pain, diarrhea, etc.
Nervous system reactions: Dizziness, headache, drowsiness, etc.
Skin reactions: Rash, itching, etc.
Respiratory system reactions: Cough, difficulty breathing, etc.
Cardiovascular system reactions: Palpitations, tachycardia, etc.
Liver reactions: Elevated transaminases, jaundice, etc.
Urinary system reactions: Frequent urination, urgency, pain during urination, etc.
Hematological system reactions: Decreased white blood cells, decreased platelets, etc.
Allergic reactions: Rash, itching, hives, angioneurotic edema, etc. ......                                                                                                                            & 1     & 1                                      \\ \hline
\textbf{Bloom with instruction-tuning:} 
The adverse reactions of acetaminophen and dihydrocodeine include but are not limited to nausea, vomiting, drowsiness, dizziness, and rashes..                                                                  & 2 & 1                                                               \\ \hline
\textbf{Bloom with knowledge-tuning:} 
Adverse reactions to acetaminophen and dihydrocodeine mainly include restlessness, nausea, vomiting, constipation, uneasiness, dizziness, excitement, and irritability. 
 &3  & 3  \\
\midrule
\midrule
\textbf{Medical Knowledge:} 
Entity: cicatricial pyloric obstruction, Attribute: symptom, Knowledge content: ``vomiting during afternoon and night, abdominal pain during the night and after eating.''                  & -         & -                               \\ \hline
\textbf{Question:} 
When should we suspect a cicatricial pyloric obstruction?                                      
 & - & -   \\ \hline

\textbf{Bloom:} 
When the following symptoms are present, suspect cicatricial pyloric obstruction: (1) Upper abdominal pain, (2) Nausea, (3) Vomiting, (4) Weight loss, (5) Belching, (6) Black stools, (7) Jaundice, ......                                                                                                                            & 2     & 2.5                                      \\ \hline
\textbf{Bloom with instruction-tuning:} 
When the following symptoms are present, suspect scar-induced pyloric obstruction: abdominal pain, vomiting, acid reflux, jaundice, heartburn, belching, nausea, loss of appetite, weight loss, .......                                                                  & 2 & 3                                                               \\ \hline
\textbf{Bloom with knowledge-tuning:} 
(1) Abdominal pain and bloating: cicatricial pyloric obstruction often occurs after eating, especially at night. The upper abdominal pain worsens after meals and progresses to diffuse upper abdominal distension or discomfort as gastric retention occurs. (2) Vomiting: The most prominent symptom, vomiting mainly occurs in the afternoon and evening. 
 & 3  & 3  \\ 
\midrule
\midrule

\textbf{Medical Knowledge:} 
Entity: Gastric cancer, Attribute: symptom, Knowledge content:``Early stages of gastric cancer: Mostly no obvious symptoms. A few might experience nausea, vomiting, or discomfort in the upper abdomen, and a feeling of fullness after eating. 
Condition progress: Symptoms such as intensified pain in the upper abdomen, decreased appetite, nausea, vomiting, fatigue, and weight loss gradually appear. Some patients may exhibit signs such as vomiting blood, passing black stools, or a lump in the upper abdomen. ''                 & -         & -                               \\ \hline
\textbf{Question:} 
What are the common symptoms of gastric cancer?                                      
 & - & -   \\ \hline

\textbf{Bloom:} 
The common symptoms of gastric cancer include:
Indigestion: Patients with gastric cancer may experience loss of appetite, indigestion, nausea, and vomiting.
Abdominal pain: Patients with gastric cancer may experience abdominal pain, bloating, and general abdominal discomfort.
Weight loss: Patients with gastric cancer may experience weight loss and thinness.
Anemia: Patients with gastric cancer may present with anemia and a pale complexion.
Jaundice: Patients with gastric cancer may develop jaundice and itchy skin.
.......                                                                                                                            & 1     & 2                                      \\ \hline
\textbf{Bloom with instruction-tuning:} 
Common symptoms of gastric cancer include upper abdominal pain, a feeling of fullness in the upper abdomen, reduced appetite, nausea, vomiting, indigestion, and black stools (tar-like stools)......                                                                  & 2 & 3                                                              \\ \hline
\textbf{Bloom with knowledge-tuning:} 
In the early stages of gastric cancer, most patients show no obvious symptoms. A few may experience nausea, vomiting, discomfort in the upper abdomen, or a feeling of fullness after eating. As the disease progresses, symptoms like increased pain in the upper abdomen, decreased appetite, nausea, vomiting, fatigue, and weight loss may emerge. Some patients might also exhibit signs such as vomiting blood, black stools, and lumps in the upper abdomen.
 & 3  & 3  \\ 
 
 \bottomrule
\end{tabularx}
\caption{Average $H_2$ score cases for the generated responses by medical specialists.}
\label{h2}
\end{table*}
\section{D. Limitations}
Knowledge-tuning enables the LLMs to formulate responses that draw upon relevant medical knowledge in the context of given questions. Despite this beneficial aspect, LLMs may still exhibit potential shortcomings, such as erroneous parameter prediction for the knowledge function or inaccuracies in effectively integrating medical knowledge into their generated responses.

\section{E. Ethics Statement}
Knowledge-tuning is primarily dedicated to research and is not intended to offer medical advice. The medical information utilized in this study is sourced from open-access medical knowledge bases. It is important to note that the accuracy of responses generated by LLMs cannot be guaranteed, and the medical knowledge utilized therein should not be construed as a substitute for professional medical advice. If one experiences any discomfort or distress, it is strongly advised to seek the guidance of a qualified medical professional.
\end{document}